\def\year{2020}\relax
\documentclass[letterpaper]{article} 
\usepackage{aaai20}  
\usepackage{times}  
\usepackage{helvet} 
\usepackage{courier}  
\usepackage[hyphens]{url}  
\usepackage{graphicx} 
\usepackage{graphicx}  
\usepackage{amssymb}
\usepackage{amsmath}
\usepackage{adjustbox}
\usepackage{multirow}
\usepackage{booktabs}
\usepackage{multirow}
\usepackage{subcaption}

\title{Sequential Recommendation with Relation-Aware Kernelized Self-Attention}
\author{Mingi Ji, Weonyoung Joo, Kyungwoo Song, Yoon-Yeong Kim, Il-Chul Moon \\
	Korea Advenced Institute of Science and Technology (KAIST), Korea \\
	\{qwertgfdcvb, es345, gtshs2, yoonyeong.kim, icmoon\}@kaist.ac.kr
}
 
\begin{document}

\maketitle

\begin{abstract}
Recent studies identified that sequential \textit{Recommendation} is improved by the \textit{attention} mechanism. By following this development, we propose Relation-Aware Kernelized Self-Attention (RKSA) adopting a self-attention mechanism of the \textit{Transformer} with augmentation of a probabilistic model. The original self-attention of Transformer is a deterministic measure without relation-awareness. Therefore, we introduce a latent space to the self-attention, and the latent space models the recommendation context from relation as a multivariate skew-normal distribution with a kernelized covariance matrix from co-occurrences, item characteristics, and user information. This work merges the self-attention of the Transformer and the sequential recommendation by adding a probabilistic model of the recommendation task specifics. We experimented RKSA over the benchmark datasets, and RKSA shows significant improvements compared to the recent baseline models. Also, RKSA were able to produce a latent space model that answers the reasons for recommendation.
\end{abstract}

\section{Introduction}
\noindent \textit{Recommendation} is one of the key application areas of artificial intelligence in the big data era. The recommendation tasks are supported by large scale data, and users need to select a specific item from many alternative items. This selection requirement motivates the utilization of attention mechanism in the recommendation task. The attention is applied to the item selection, and the sequential recommendation particularly selects the past item choice records to consider the recommendation at the current timestep with the attention mechanism \cite{atem,stamp,narm,shan,marank,csan}. 

Given the relationship between the attention and the recommendation, adopting a new attention mechanism to the recommendation has been a research trend. For instance, Self-Attentive Sequential Recommendation (SASRec) \cite{sasrec} adopted the self-attention mechanism of the Transformer \cite{transformer} to the recommendation task. This adaptation is interesting, but it was limitedly customized to meet the task specifics. Recommendation often requires understanding items, users, browsing sequences, etc, and the recommendation models need to consider such contexts which SASRec does not provide. Following SASRec, there have been developments in using the self-attention mechanism of the Transformer to model a task specific feature of sequential recommendation. For example, ATRank \cite{atrank} utilized the self-attention mechanism for considering the influences from heterogenous behavior representations. To model the user's short-term intent, AttRec \cite{attrec} adopted the self-attention mechanism on the user interaction history. Similar to ATRank and AttRec, BST \cite{bst} used the self-attention mechanism for aggregate of the auxiliary user and item features. 
\begin{figure}
	\includegraphics[width=\columnwidth]{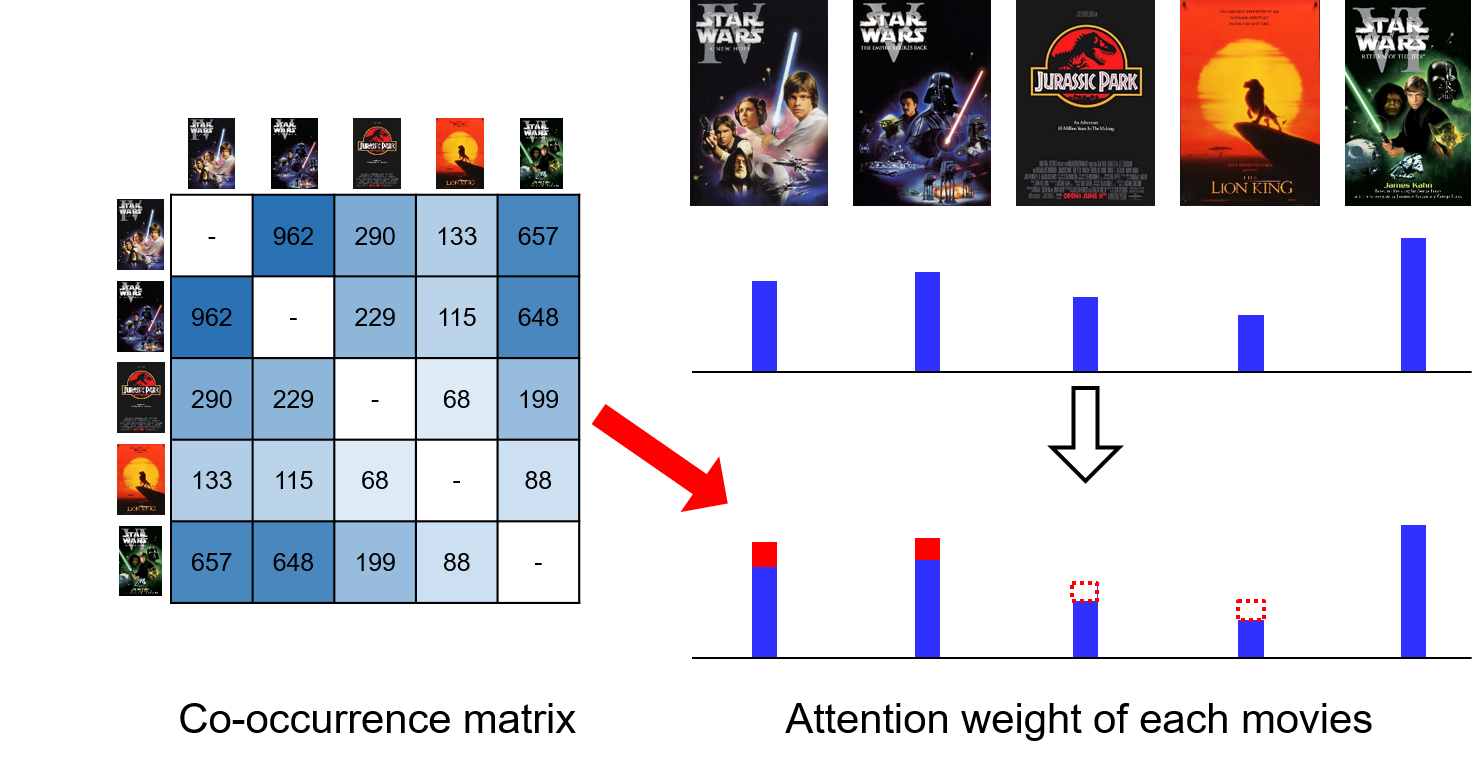}
	\caption{Each entry of the co-occurrence matrix means the number of users that each movie pair appeared together in an user sequence in the MovieLens dataset. We can see that there are many users who watched \textit{Star Wars} movies together. This allows modifying the attention weight from blue to red using co-occurrence information, when \textit{Star Wars 6} is a query.}
	\label{fig:example_v1}
\end{figure}

Given the success of the self-attention \cite{self-att1,self-att2,self-att3}, the recommendation task can be improved from the sequential information, which was limitedly used in the previous works. Moreover, such utilization on the sequential information provides a new approach to customize the self-attention structure to the recommendation task. Figure \ref{fig:example_v1} is the example that the co-occurrence information may influence the attention weight. It is feasible to see a movie pair that has a higher co-occurrence than others, and this movie pair should inform the attention mechanism to increase the weight. 

We renovate and customize the self-attention of Transformer with a latent space model. Specifically, we add a latent space to the self-attention value of the Transformer, and we use the latent space to model the context from relations of the recommendation task. The latent space is modeled as a multivariate skew-normal (MSN) distribution \cite{msn} with the dimension of the number of unique items in the sequence. The covariance matrix of the MSN distribution is the variable that we model the relations of a sequence, items, and a user by a kernel function that provides the flexibility of the recommendation task adaptation. After the kernel modeling, we provide the reparametrization of the MSN distribution to enable the amortized inference on the introduced latent space. Since the relation modeling is done with kernelization, we call this model as relation-aware kernelized self-attention (RKSA). 
We designed RKSA with three innovations. First, the deterministic Transformer may not work well in the generalized task of recommendation because of sparsity, so we added a latent dimension and its corresponding reparameterization. Second, the covariance modeling with the relation-aware kernel enables the more fundamental adaptation of the self-attention to the recommendation. Third, the kernelized latent space of the self-attention provides the reasoning on the recommendation result. RKSA is evaluated against eight baseline models including SASRec, HCRNN, NARM, etc; as well as, five benchmark datasets with Amazon review, MovieLens, Steam, etc. Our experiments showed that RKSA significantly improves the performance over the baselines on the benchmarks, consistently. 

\section{Preliminary}
\subsubsection{Multi-Head Attention}
We start the preliminary by reviewing the self-attention structure that is the backbone of RKSA. Recently, \cite{transformer} proposed the \textit{scaled-dot product attention}, which is defined by Equation \ref{scaled-dot_product_attention} where $Q \in \mathbb{R}^{n \times d_k}$, $K \in \mathbb{R}^{m \times d_k}$, and $V\in \mathbb{R}^{m \times d_v}$ are the queries, the keys, and the value matrix, respectively.
The scaled-dot product attention calculates importance weights from the dot-product of query $i$ with key $j$ with a scaling of $\sqrt{d_k}$. This importance is boundarized by the softmax, and the boundarized importance is again multiplied by the value $v$ to form the scaled-dot product attention. 
\begin{equation}\label{scaled-dot_product_attention}
\text{Attention}(Q,K,V) = \text{softmax}(\frac{QK^T}{\sqrt{d_k}})V
\end{equation}

When the query, the key, and the value take the same $X \in \mathbb{R}^{n \times d}$ as an input matrix in Equation \ref{self-attention}, the scaled-dot product attention is called as the \textit{self-attention}.
A self-attention with an additional predefined or learnable \textit{positional embedding} \cite{transformer,sasrec} is able to capture the latent information of the position like previous recurrent networks.
\begin{equation}\label{self-attention}
\text{SA}(X) = \text{Attention} (XW^Q,XW^K,XW^V)
\end{equation}

\textit{Multi-head attention} uses $H$ scaled-dot product attentions with $1/H$ times smaller dimension on attention weight parameters. \cite{transformer} found that the multi-head attention is useful even though it uses the similar number of parameters compared to the single-head attention.
\begin{align} \label{mha}
\text{MHA}(Q,K,V) = [\text{Head}_1 , ... , \text{Head}_H] W^O  \nonumber \\
\text{where} ~~ \text{Head}_i = \text{Attention}(XW_i^Q,XW_i^K,XW_i^V)
\end{align}
\cite{casan} considered the dependencies, i.e. item co-occurrence, between the temporal state representations over a single sequence with the scaled-dot product attention. Their model is introducing a context vector $C$ to be linearly combined with $Q$ and $K$ in the self-attention. We expand this context modeling with stochasticity and kernel method to add the flexibility of the self-attention. 

\subsubsection{Multivariate Skew-Normal Distribution}
As we mentioned the latent space model of RKSA, we introduce an explicit probability density model to the self-attention structure. Here, we choose the \textit{multivariate skew-normal} (MSN) distribution to be the explicit density because we intend to model 1) the covariance structure between items; and 2) the skewness of the attention value. It would be natural to consider the multivariate normal distribution to enable the covariance model, but the normal distribution is unable to model the skewness because it enforces the symmetric shape of the density curve. As the name suggests, the MSN distribution reflects the skewness as the shape parameter $\alpha$ \cite{msn}. The MSN distribution needs four parameters: location $\xi$, scale $\omega$, correlation $\psi$, and shape $\alpha$. Following \cite{msn-app}, a $k$-dimensional random variable $x \in \mathbb{R}^k$ follows the MSN disitribution with the location parameter $\xi\in\mathbb{R}^k$; the correlation matrix $\psi\in\mathbb{R}^{k \times k}$; the scale parameter $\omega=\text{diag}(\omega_1,...,\omega_k)\in\mathbb{R}^{k\times k}$; and the shape parameter $\alpha\in\mathbb{R}^k$, as Equation \ref{MSN pdf}.
\begin{equation}\label{MSN pdf}
f(x)=2\phi_k(x;\xi,\Sigma)\Phi(\alpha^T\omega^{-1}(x-\xi))
\end{equation}
Here, $\Sigma=\omega\psi\omega$ is the covariance matrix; $\phi_k$ is the $k$-dimensional multivariate normal density with the mean $\xi$ and the covariance $\Sigma$; and $\Phi$ is the cumulative distribution function of $N(0,1)$. If $\alpha$ is a zero vector, the distribution reduces down to the multivariate normal distribution with the mean $\xi$ and the covariance $\Sigma$.

\begin{figure*}
	\begin{center}
	\includegraphics[width=1.8\columnwidth]{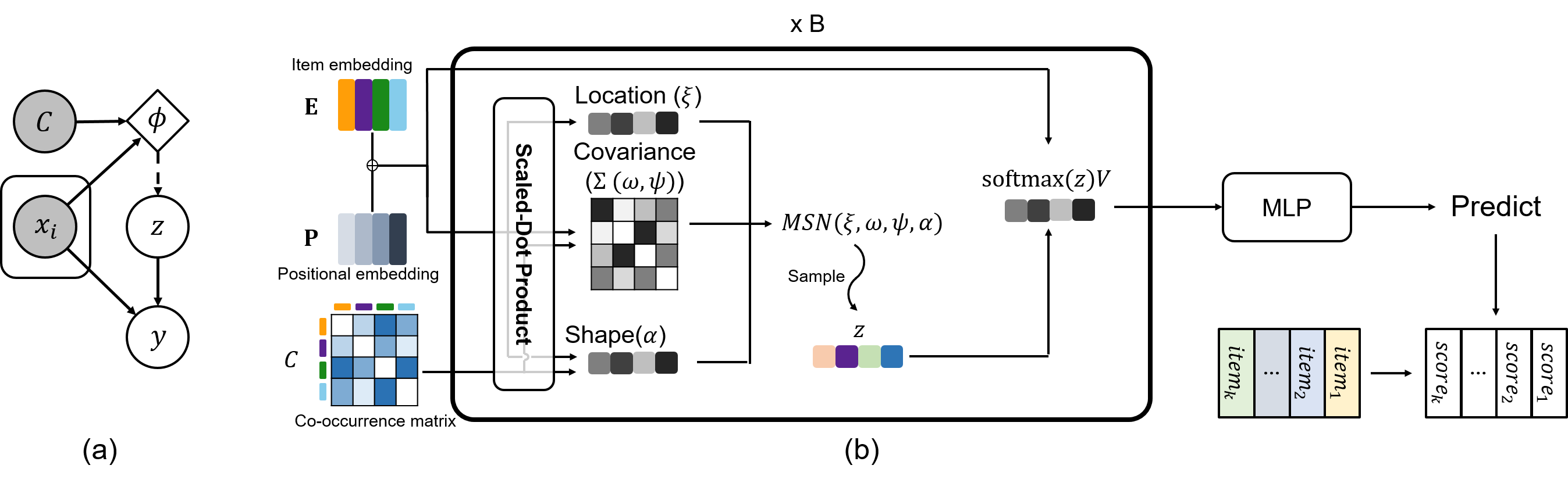}
	\caption{(a) Graphical notation of RKSA. $\phi$ is the parameter of MSN distribution, and dashed line denotes sampling procedure. (b) The overall structure of RKSA with MSN parameters. The scaled-dot product denotes the matrix multiplication between query and key matrix in scaled-dot product attention.}
	\label{fig:graphical_v2}
	\end{center}
\end{figure*}

\subsubsection{Kernel Function}
Given that we intend to model the covariance of the MSN, we introduce how we provide the flexible covariance structure through kernels. Kernel function, $k(x,x') = \phi(x) \cdot \phi(x')$, evaluates a pair of observations in the observation space $\mathcal{X}$ with a real value.
In the machine learning field, the kernel functions are widely used to compute the similarity between two data points as a covariance matrix.
Given observations $\mathcal{X} = \{x_i\}_{i=1}^n$, a function $k: \mathcal{X} \times \mathcal{X} \rightarrow \mathbb{R}^k$ is a valid kernel if and only if it is (1) symmetric: $k(x,x') = k(x',x)$ for all $x,x' \in \mathcal{X}$; and (2) positive semi-definite: $\sum_{i,j} a_i a_j k(x_i,x_j) \geq 0$ for all $a_i,a_j \in \mathbb{R}$ \cite{kernel}. We apply a customized kernel function to model the relational covariance parameter of the MSN in RKSA, and we provide proofs on the validity of our customized kernels.

\section{Methodology}
This section explains the sequential recommendation task, the overall structure of Relation-Aware Kernelized Self Attention (RKSA), and its detailed parameter modeling.

\subsection{Problem Statement}
A sequential recommendation uses datasets built upon a past action sequence of a user. Let $\mathcal{U}=\{u_1,u_2,...,u_{|\mathcal{U}|}\}$ be a set of users; let $\mathcal{I}=\{i_1,i_2,...,i_{|\mathcal{I}|}\}$ be a set of items; and let $\mathcal{S}_u=\{i_1^{(u)},i_2^{(u)},...,i_{n_u}^{(u)}\}$ be a user $u$'s action sequence. The task of sequential recommendation is predicting the next item to interact by the user, as $P(i_{n_u+1}^{(u)}=i|\cup_{u \in \mathcal{U}} \mathcal{S}_u)$. 

\subsection{Self-Attention Block}
We propose elation-aware Kernelized Self-Attention (RKSA), which is a modification of the self-attention structure embedded in Transformer \cite{transformer}. Figure \ref{fig:graphical_v2} illustrates that RKSA is a customized self-attention based on relations, such as the item, the user, and the global co-occurrence information. The detailed procedure is explained in the below.  

\subsubsection{Embedding Layer}
Since the raw data of items and interactions follow sparse one-hot encoding, we need to embed the information of items and positions of interactions. To create such embeddings, we use the latest $n$ actions from the user sequence of $\mathcal{S}_u$. Specifically, the item embedding matrix is defined as $\textbf{E}\in\mathbb{R}^{|\mathcal{I}|\times d}$, where $d$ denotes the dimensionality of the embedding. $\textbf{E}$ is estimated by a hidden layer as a part of the modeled neural network, and the raw input to the hidden layer is the one-hot encoding of an interacted item at time $t$. Similarly, we set a user embedding to be $\textbf{\text{U}} \in \mathbb{R}^{ |\mathcal{U}| \times d}$ to make a distinction between users. Also, we define a positional embedding matrix as $\textbf{P}\in\mathbb{R}^{n \times d}$, to introduce the sequential ordering information of the interactions, which we follow the ideas from \cite{sasrec}. $\textbf{P}$ and $\textbf{U}$ are also estimated by a hidden layer that matches the dimensionality of $\textbf{E}$ for the further construction of $x_t$.

Afterward, we estimate the inputs to RKSA, and the input should convey the representation of items and positions in the sequences. Here, we assume that the item at time $t$, which is $i_t$, is represented as $x_t$ as a timestep of the sequence, and we denoted the representation as $x_t$ because it is the input to RKSA. $x_t$ is estimated through the summation of the item embedding $e_{i_t}\in\textbf{E}$ and the positional embedding $p_t\in\textbf{P}$, as $x_t=e_{i_t} + p_t$. Finally, the input item sequence is expressed as $X\in\mathbb{R}^{n \times d}$ by combining the item embedding $\textbf{E}$, and the positional embedding $\textbf{P}$. 

\subsubsection{Relation-Aware Kernelized Self-Attention}
The core component of RKSA is the multi-head attention structure that includes a latent variable of $z$. Given that Equation \ref{scaled-dot_product_attention} is deterministic, we intend to turn $\frac{QK^T}{\sqrt{d_k}}$ into a single latent variable $z$. The changed part is originally the alignment score of the attention mechanism, so its range becomes $\mathbb{R}$. Additionally, we assume that there is a skewed shape in the alignment score distribution, so we designed $z$ to follow the multivariate skew-normal distribution (MSN), as Equation \ref{att}. In other words, we sample the logit of the softmax function from the MSN distribution.
\begin{equation}\label{att}
\begin{split}
	\textbf{\text{H}}=\text{RKSA}(X,C)=\text{softmax}(Z)V \\
	\text{where} ~~ z \sim MSN(Z|\xi,\Sigma,\alpha)
\end{split}
\end{equation}

In the above, the parameters of the MSN distribution include the location $\xi$, the covariance $\Sigma$, and the shape parameter $\alpha$. The details of the parameters are explained in Section \textit{Parameter Modeling}.
Additionally, in Equation \ref{att}, $X$ denotes the items in the sequence $\mathcal{S}_{u}$, and $C$ is the co-occurrence matrix of $\mathcal{S}_{u}$ from our kernel model, which is explained in Section \textit{Covariance}. The co-occurrence matrix $C$ is constructed by counting the co-occurrence number between item pairs in the whole dataset. We follow the amortized inference with a reparametrization on the MSN; and $X$ and $C$ are used as inputs to the inference. 

Lastly, the output of RKSA is the hidden dimension defined as \textbf{H}$=\{h_1,h_2,...,h_n\}, h_i \in R^d$ in Equation \ref{att}. $V$ is the value vector estimated from the input item sequence representation $X$. Since we modify the scaled-dot product attention, RKSA is easily expanded to be a variant of multi-head attention by following the same procedure of Equation \ref{mha}.

\subsubsection{Point-Wise Feed-Forward Network}
We apply the \textit{Point-Wise Feed-Forward Network} in Transformer to the output of RKSA by each position. The point-wise feed-forward network consists of two linear transformations with a ReLU nonlinear activation function between the linear transformations. 
The final output of the point-wise feed-forward network, $F$ is $\{\text{FFN}(h_1),...,\text{FFN}(h_d)\}$.

Besides of the above modeling structure, we stacked multiple self-attention blocks to learn complex transition patterns, and we added residual connections \cite{resnet} to train a deeper network structure. We also applied the layer normalization \cite{layernorm} and the dropout \cite{dropout} to the output of each layer by following \cite{transformer}.

\subsubsection{Output Layer}
Let $B$ be the number of self-attention blocks. The task requires predicting the ($n+1$)-th item with the $n$-th output of the $B$-th self-attention block. We use the same weights of the item embedding layer to rank the item prediction. The relevance score of the item $i_n$ is defined as $r_{i,n}$:
\begin{equation}\label{rank}
r_{i,n}=F^{(B)}_{n}\textbf{\text{E}}_i
\end{equation}
$F^{(B)}_{n}$ denotes the $n$-th output of the last self-attention block, and $\textbf{\text{E}}_i$ is the embedding of item $i$. The prediction ranking of the ($n+1$)-th item is defined by the ranking of the items' relevance scores.

\subsection{Parameter Modeling}
This section enumerates the detailed modeling of the MSN parameters, which is used for the latent variable $z$ in RKSA.

\subsubsection{Location}
The location $\xi$ has the same role of the mean of multivariate normal distribution. Given that we use the MSN to sample the alignment score, we still need to provide the deterministic alignment score with the most likelihood. Therefore, we allow the alignment score to be the location parameter as:
\begin{equation}\label{location}
\xi=f(\frac{(XW_l^Q)(XW_l^K)}{\sqrt{d}})
\end{equation}
Also, we can use activation function $f$ and scaling to $\xi$ with $\sqrt{d}$. 

\subsubsection{Covariance}
The covariance $\Sigma$ represents the relation between items. While $\Sigma$ is a square matrix of parameters, $\Sigma$ has a limited size because we only use the latest $n$ items; and because there are not many unique items in those latest interactions. The relation can be measured by various methods, ranging from a simple co-occurrence counting to a non-linear kernel function. This paper design a kernel function to measure the relation between a pair of items because the kernel function is known to be the efficient and nonlinear high-dimensional distance metric that can also be learned through optimizing the kernel hyperparameters.

We compose a kernel function by considering the relations of the co-occurrence, the item and the user. For a given sequence, for timesteps $i$ and $j$, we utilize the normalized representations of $\hat{x}_i$ and $\hat{x}_j$. Additionally, we infer the variance $\omega_i^2, \omega_j^2\in\mathbb{R}_+$ of $z$ at timestep $i$ and $j$, by an amortized inference as Equation \ref{variance}. 
\begin{equation}\label{variance}
\omega_i = \text{softplus}(\frac{(x_n W_\omega^Q)(x_i W_\omega^K)}{\sqrt{d}})
\end{equation}
In the above, we set the activation function of standard deviation as softplus to make the value of standard deviation positive. The following defines three different kernel functions, and we denote $\hat{x}_i$ as $x_i$ for simplicity.

\begin{itemize}
	\item \textbf{Counting kernel} is defined by the co-occurrence number of each item pair. The counting kernel is $k_c (x_i,x_j) = \omega_i \omega_j \frac{P_{ij} ^2}{P_i P_j}$ where $P_i, P_j$ are the number of occurrence of item $i$ and $j$, respectively, and $P_{ij}$ is the number of co-occurrence of item $i$ and $j$. 
	\item \textbf{Item kernel} utilizes the representation of each item. There are two alternative kernels. The linear item kernel is $k_i (x_i,x_j) = \omega_i \omega_j (x_i \cdot x_j)$ where $\cdot$ denotes dot product; and the Radial Basis Function (RBF) kernel is $k_i(x_i,x_j)=\omega_i\omega_j\exp(-||x_i-x_j||^2)$.
	\item \textbf{User kernel} utilizes the representation of each items and users. The user kernel is $k_u (x_i,x_j) = \omega_i \omega_j [(W_s u_s \odot x_i) \cdot (W_s u_s \odot x_j)]$ for user embedding $u_s \in \mathbb{R}^d$ and weight matrix $W_s \in \mathbb{R}^{d \times d}$ where $\odot$ denotes Hadamard product.
\end{itemize}

Unlike the item and the user kernel, the validity of the counting kernel should be checked because it is not a well-known format as the linear or the RBF kernels. The counting kernel is always symmetric and positive semi-definite. Therefore, the counting kernel is a valid kernel function. 

From the property of kernel functions, we combine kernel functions by their summation to make the final kernel function flexible. The final kernel function is defined as:
\begin{gather}\label{kernel}
k(x_i,x_j)=r_1 k_c(x_i,x_j) + r_2 k_i(x_i,x_j) + r_3 k_u (x_i,x_j) \nonumber \\
\text{where} ~~  \textbf{r} = \text{softmax}(uW_u + b_u)
\end{gather}
With the above kernel function, our modeling on the correlation matrix is $\psi_{i,j}=\frac{k(x_i,x_j)}{\omega_i \omega_j}$, similar to the definition of $\Sigma$ of Equation \ref{MSN pdf}.

This section describes the covariance modeling with the final kernel, so the kernel hyperparameter, such as $\omega$, $W_s$ and $\textbf{r}$, needs to be inferred. While they need to be supervised to learn the kernel hyperparameters, the loss of the recommendation task needs to be augmented with an additional loss to guide the kernel hyperparameter. Therefore, we modeled a loss that regularizes the covariance to be the item co-occurrence. Since we have other loss terms, i.e. the recommendation loss, the learned correlation does not become same to the item co-occurrence, but the co-occurrence loss can be prior knowledge. Particularly, we measure the co-occurrence loss $L_{rank}$ with the listwise ranking loss to match the alignment of the correlation and the ranking of the item co-occurrences. The co-occurrence loss is defined as maximizing te listwise ranking loss \cite{listmle}.


\subsubsection{Shape}
The shape parameter $\alpha$ reflects the relation between a final item and an item in a user sequence. We designate $\alpha=\{\alpha_1,...,\alpha_n\}$ to items $\{i_1,...,i_n\}$. We define $\alpha$ by introducing a ratio parameter $\hat{\alpha}$ with the co-occurrence matrix $C$; and a learnable scaling parameter $s$. Specifically, we assume $\alpha_j = s_j \frac{\hat{\alpha}_j}{\text{max}(\hat{\alpha})}$, which is a scaled correlation between the final item $i_n$ and the item $i_j$.

First, we calculate the ratio parameter $\hat{\alpha}_j\in[0,1]$ with the co-occurrence matrix $C$, by the summation of the linear alignments between the last time $i_n$, and the aligned item $i_j$. Here, let $c_{i,j}$ be the value of $i$-th row and $j$-th column of co-occurrence matrix, $C$. For simplicity, we denote $c_{i_j,i_k}$ as $c_{j,k}$. The following is the detailed formula of $\hat{\alpha}_j$.
\begin{align}\label{alpha}
\hat{\alpha}_{j} & = \sum_{k=1}^{n} c_{j,k} c_{k,n} \text{~~where~} c_{k,k} \leftarrow \frac{\sum_{l \in \{1,\cdots,n\} \backslash k} c_{k,l}}{n-1} \nonumber \\
& = \sum_{k=\{1,\cdots,n\} \backslash\ \{j,n\}} c_{j,k} c_{k,n} \nonumber \\
& + \Big(\frac{\sum_{l \in \{1,\cdots,n\} \backslash j} c_{j,l}}{n-1} \Big) c_{j,n} + c_{j,n} \Big(\frac{\sum_{k=1}^{n-1}c_{k,n}}{n-1} \Big) 
\end{align}
Equation \ref{alpha} computes $\hat{\alpha}_j$ by the dot-product between $j$-th row and $n$-th column of the co-occurrence matrix $C$, which means that we calculate the correlation between the co-occurrence of $i_j$ and $i_n$. Having said that, the co-occurrence of the same item is semantically meaningless in $C$, so such cases used the average of the remaining elements in each row in the dot-product process. $\hat{\alpha}_j$ enables modeling the two-hop dependency between $i_j$ and $i_n$ through $i_k$. 

Second, Equation \ref{scaling_parameter} defines the scaling parameter, $s_j$:
\begin{equation} \label{scaling_parameter}
s_j = f(\frac{(x_n W_s^Q)(x_j W_s^K)}{\sqrt{d}})
\end{equation}
We can apply the softplus activation to $f$, so the shape parameter becomes positive. 

\subsection{Model Inference}
\subsubsection{Loss Function}
Given the above model structure, this subsection introduces the inference on the latent variable $z$ following the MSN distribution. It is well-known that the latent variable can be inferred by optimizing the evidence lower bound from the Jensen's inequality, so we optimize the evidence lower bound on the marginal log-likelihood, $p(y_n)$, when predicting the $(n+1)$-th item $i_n$. Equation \ref{prediction_loss} describes the loss function of this prediction task.
\begin{align}\label{prediction_loss}
L_z&=\mathbb{E}_z[\log p(y_n|z)] =\int\log p(y_n|z)p(z)dz  \\
&\leq \log\int p(y_n|z)p(z)dz =\log p(y_n) \nonumber
\end{align}

$L_z$ utilizes the binary cross-entropy loss with the negative sampling as conducted in \cite{sasrec} to calculate $p(y_n|z)$. 
It should be noted that the actual loss function is a combination of the prediction loss and the co-occurrence loss, which is $L=L_z+\lambda_r L_{rank}$. $\lambda_r$ is the regularization weight hyperparameter of the co-occurrence loss. 

\subsubsection{Reparametrization of $Z$} 
We sample the values of $z$ from the $MSN(Z|\xi,\omega,\psi,\alpha)$ distribution using the reparameterization trick. Equation \ref{reparametrization} shows the reparametrization of the MSN distribution with the sample from the two Normal distributions. 
\begin{align} \label{reparametrization}
& y_0 \sim N(Y_0|0,1), ~~ y \sim N(Y|0, \psi), ~~ \delta_j=\frac{\alpha_j}{\sqrt{1+\alpha_j^2}}  \\
& \hat{z}_j=\delta_j|y_0|+(1-\delta_j^2)^{\frac{1}{2}}y_j, ~~ z_j=\xi_j+\omega_j\hat{z}_j \nonumber 
\end{align}

This reparametrization is utilized because $z$ needs to be instantiated for the forward path. Equation \ref{reparametrization} shows how to sample $z$ given the amortized inference parameters of $\alpha$, $\xi$, $\omega$, and $\psi$. Once the forward path is enabled, the neural network can be trained via the back-propagation method.

\section{Experiment Result}

\subsubsection{Datasets}
We evaluate our model on five real world datasets: Amazon (Beauty, Games) \cite{amazon-1,amazon-2}, CiteULike, Steam, and MovieLens. We follow the same preprocessing procedure on Beauty, Games, and Steam from \cite{sasrec}. For preprocessing CiteULike and MovieLens, we follow the preprocessing procedure from \cite{hcrnn}. We split all datasets for training, validation, and testing following the procedure of \cite{sasrec}. Table \ref{dataset} summarizes the statistics of the preprocessed datasets.
\begin{table}
	\centering
	\caption{Statistics of evaluation datasets.}
	\label{dataset}
	\begin{adjustbox}{max width=\columnwidth}
		\begin{tabular}{l r r r r r}
			\toprule
			\multirow{3}{*}{Dataset} & \multirow{3}{*}{\#users} & \multirow{3}{*}{\#items} & \multirow{3}{*}{\#actions} & \multicolumn{1}{c}{avg.} & \multicolumn{1}{c}{avg.} \\
			& & & & \multicolumn{1}{c}{actions} & \multicolumn{1}{c}{actions}\\ 
			& & & & \multicolumn{1}{c}{/user}& \multicolumn{1}{c}{/item}\\
			\midrule
			Beauty & 52,024 & 57,289 & 0.4m & 7.6 & 6.9 \\
			Games & 31,013 & 23,715 & 0.3m & 9.3 & 12.1 \\
			CiteULike & 1,798 & 2,000 & 0.05m & 30.6 & 27.5 \\
			Steam & 334,730 & 13,047 & 3.7m & 11.0 & 282.5 \\
			MovieLens & 4,639 & 930 & 0.2m & 40.9 & 204.0 \\
			\bottomrule
		\end{tabular}
	\end{adjustbox}
\end{table}

\begin{table*}[]
	\centering
	\caption{Performance comparison (higher is better). The best performing model is indicated as boldface. The second-best model is indicated as underline. $*$ indicates that the result has p-value less than $0.05$ against the second-best result based on t-test.
		\label{tab:result}} %
	\begin{adjustbox}{max width=2\columnwidth}
		\begin{tabular}{clccccccccc}
			\toprule
			Dataset                    & Metric  & Pop    & Item-KNN & BPR-MF  & GRU4REC & NARM   & HCRNN & AttRec & SASRec & RKSA   \\
			\midrule
			\multirow{4}{*}{Beauty}    & Hit@5   & 0.2972 & 0.0885  & 0.0735 & 0.3097 & 0.3663 & 0.3643 & 0.3341 & \underline{0.3735} & \textbf{0.3999*} \\
			& NDCG@5  & 0.1478 & 0.0872  & 0.0486 & 0.2257 & 0.2785 & 0.2764 & 0.2535 & \underline{0.2846} & \textbf{0.2998*} \\
			& Hit@10  & 0.4289 & 0.0885  & 0.1285 & 0.4174 & 0.4674 & 0.4653 & 0.4222 & \underline{0.4720} & \textbf{0.5015*} \\
			& NDCG@10 & 0.1882 & 0.0872  & 0.0662 & 0.2604 & 0.3111 & 0.3091 & 0.2819 & \underline{0.3164} & \textbf{0.3326*} \\
			\midrule
			\multirow{4}{*}{Games}     & Hit@5   & 0.3416 & 0.1969  & 0.1291 & 0.5749 & 0.6224 & 0.6229 & 0.5673 & \underline{0.6395} & \textbf{0.6544*} \\
			& NDCG@5  & 0.1730 & 0.1892  & 0.0920 & 0.4570  & 0.4927 & 0.4955 & 0.4358 & \underline{0.5068} & \textbf{0.5168*} \\
			& Hit@10  & 0.4846 & 0.1969  & 0.1919 & 0.6733  & 0.7244 & 0.7233 & 0.6812 & \underline{0.7373} & \textbf{0.7551*} \\
			& NDCG@10 & 0.2168 & 0.1892  & 0.1121 & 0.4889 & 0.5257 & 0.5281 & 0.4727 & \underline{0.5385} & \textbf{0.5495*} \\
			\midrule
			\multirow{4}{*}{CiteULike} & Hit@5   & 0.1318 & 0.3563  & 0.1624 & 0.4310  & 0.4457 & 0.4442 & 0.4275 & \underline{0.5044} & \textbf{0.5308} \\
			& NDCG@5  & 0.0650 & 0.2666  & 0.1107 & 0.2982 & 0.3016 & 0.3053 & 0.2891 & \underline{0.3447} & \textbf{0.3687*} \\
			& Hit@10  & 0.2144 & 0.3815  & 0.2472 & 0.5879 & 0.6150  & 0.6077 & 0.5808 & \underline{0.6757} & \textbf{0.6893*} \\
			& NDCG@10 & 0.0902 & 0.2751  & 0.1378 & 0.3488 & 0.3565 & 0.3583 & 0.3388 & \underline{0.4001} & \textbf{0.4202*}\\
			\midrule
			\multirow{4}{*}{Steam}     & Hit@5   & 0.5545 & 0.2964  & 0.5724 & 0.7065 & 0.7095 & 0.7136 & 0.5936 & \underline{0.7477} & \textbf{0.7514} \\
			& NDCG@5  & 0.2873 & 0.2724  & 0.4144 & 0.5444 & 0.5476 & 0.5516 & 0.4182 & \underline{0.5828} & \textbf{0.5841} \\
			& Hit@10  & 0.7162 & 0.2965  & 0.7083 & 0.8293 & 0.8314 & 0.8344 & 0.7491 & \underline{0.8610} & \textbf{0.8668*} \\
			& NDCG@10 & 0.3370 & 0.2724  & 0.4587 & 0.5844 & 0.5873 & 0.5909 & 0.4687 & \underline{0.6196} & \textbf{0.6217} \\
			\midrule
			\multirow{4}{*}{MovieLens}     & Hit@5   & 0.1521 & 0.2950  & 0.1241 & 0.3883 & 0.4057 & 0.4039 & 0.3493 & \underline{0.4260} & \textbf{0.4361*} \\
			& NDCG@5  & 0.0733 & 0.2019  & 0.0767 & 0.2650  & 0.2775 & 0.2770 & 0.2217 & \underline{0.2965} & \textbf{0.3023*} \\
			& Hit@10  & 0.2547 & 0.4051  & 0.2088 & 0.5487 & 0.5617 & 0.5606 & 0.5094 & \underline{0.5873} & \textbf{0.5997*} \\
			& NDCG@10 & 0.1044 & 0.2376  & 0.1039 & 0.3167 & 0.3278 & 0.3275 & 0.2734 & \underline{0.3485} & \textbf{0.3552*} \\
			\bottomrule
		\end{tabular}
	\end{adjustbox}
\end{table*}

\begin{table}[]
	\caption{Ablation study on the Beauty and MovieLens datasets. The measure is Hit@10 and C, I and U denote counting, item and user kernel function respectively. B is the Beauty dataset; and M is the MovieLens dataset.}
	\label{kerenl-ab}
	\begin{tabular}{lccccccc}
		\toprule
		\textbf{} & \small{C}      & \small{I}      & \small{U}      & \small{C+I}  & \small{C+U}  & \small{I+U}  & \small{C+I+U} \\
		\midrule
		\small{B}    & \textbf{\scriptsize{0.5015}} & \scriptsize{0.4982} & \scriptsize{0.4958} & \scriptsize{0.5012} & \scriptsize{0.4955} & \scriptsize{0.4951} & \scriptsize{0.5011}    \\
		\scriptsize{M} & \scriptsize{0.5911} & \scriptsize{0.5966} & \scriptsize{0.5977} & \scriptsize{0.5960}  & \scriptsize{0.5962} & \textbf{\scriptsize{0.5997}} & \scriptsize{0.5973} \\
		\bottomrule
	\end{tabular}
\end{table}

\subsubsection{Baselines}
We compared RKSA with seven baselines. 
\begin{itemize}
	\item \textbf{Pop} always recommends the most popular items.
	\item \textbf{Item-KNN} \cite{item-knn} recommends an item based on the measured similarity of the last item.
	\item \textbf{BPR-MF} \cite{bprmf} recommends an item by the user and the item latent vectors with the matrix factorization. 
	\item \textbf{GRU4REC} \cite{gru4rec} models the sequential user history with GRU and the specialized recommendation loss function such as Top1 and BPR loss.
	\item \textbf{NARM} \cite{narm} focuses on both short and long-term dependency of a sequence with an attention and a modified bi-linear embedding function. 
	\item \textbf{HCRNN} \cite{hcrnn} considers the user's sequential interest change with the global, the local, and the temporary context modeling. It modifies the GRU cell structure to incorporate the various context modeling. 
	\item \textbf{AttRec} \cite{attrec} models the short-term intent using self-attention and the long-term preference with metric learning.
	\item \textbf{SASRec} \cite{sasrec} is a Transformer model which combines the strength of Markov chains and RNN. SASRec focuses on finding the relevant items adaptively with self-attention mechanisms.
\end{itemize}

\subsubsection{Experiment Settings}
For GRU4REC, NARM, HCRNN, and SASRec, we use the official codes written by the corresponding authors. For GRU4REC, NARM and HCRNN, we apply the data augmentation method proposed by NARM \cite{narm}. We use two self-attention blocks and one head for SASRec and RKSA following the default setting of \cite{sasrec}. For fair comparisons, we apply the same setting of the batch size (128), the item embedding (64), the dropout rate (0.5), the learning rate (0.001), and the optimizer (Adam). We use the same setting of authors for other hyperparameters. For RKSA, we set the co-occurrence loss weight $\lambda_r$ as 0.001. Furthermore, we use the learning rate decay and the early stopping based on the validation accuracy for all methods. We use the latest 50 actions of sequence for all datasets. 

\subsection{Quantitative Analysis}
Table \ref{tab:result} presents the recommendation performance of the experimented models. We adopt two widely used measurements: Hit Rate@K and NDCG@K \cite{measure}. Considering that all user-item pairs require heavy computation, we use 100 negative samples for the evaluation following \cite{sasrec,measure}. We repeat each experiment for five times, and the results are the average of each method. The performance of RKSA comes from the best kernel variant of RKSA, and RKSA outperforms all baseline models on all datasets and metrics. Especially, Beauty shows the biggest improvement. Beauty is the most sparse dataset, so there are many items infrequently occurred. This result suggests that using the relational information can be helpful for predicting such infrequent items.

\subsubsection{Ablation Study}
We compared the kernel function combinations on Beauty and MovieLens datasets. We consider Beauty as a representative sparse dataset, and MovieLens as a representative dense dataset. Table \ref{kerenl-ab} shows the performance of each kernel functions. We assume that using the sparse and short dataset is hard to learn the representation of the item and user. Therefore, RKSA with the counting kernel function shows the best performance on the sparse dataset. On the contrary, it is relatively easy to learn the representation of item and user by the dense dataset, and Table \ref{kerenl-ab} shows the kernel combination of the item and the user is best.

\subsection{Qualitative Analysis}

\subsubsection{Item Embedding and Correlation Matrix}
\begin{figure}
	\includegraphics[width=\columnwidth]{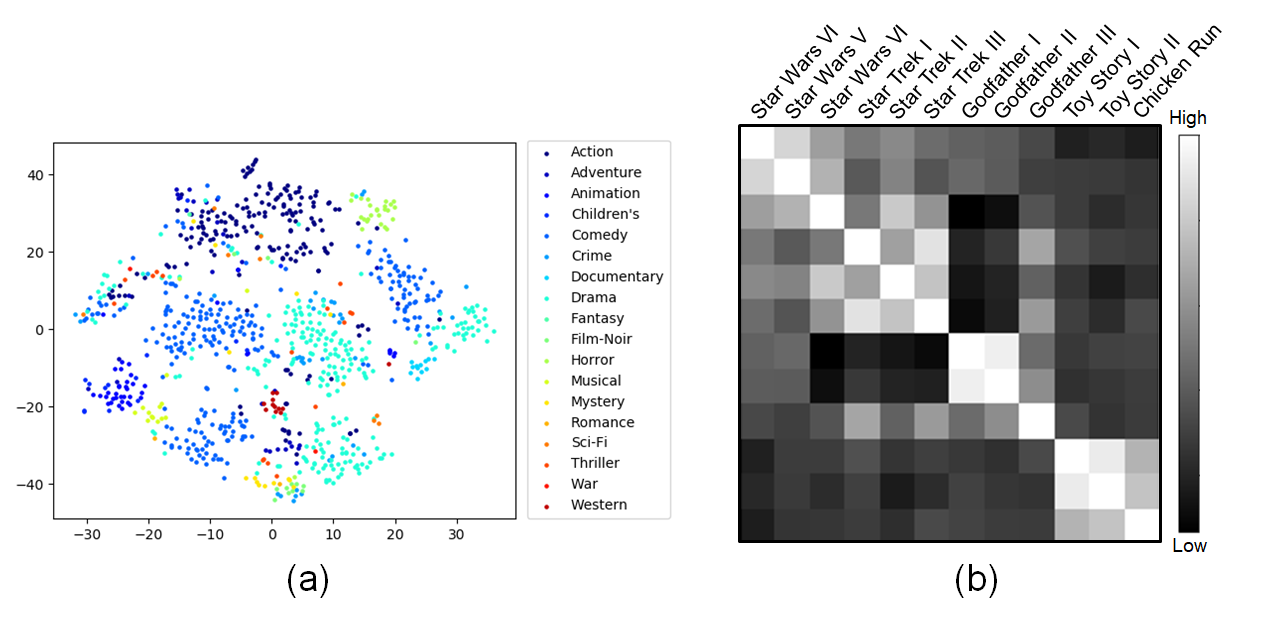}
	\caption{(a) Item embedding visualization with tSNE \cite{tsne} of MovieLens dataset. (b) Correlation between movies by the counting and item kernel combination.}
	\label{item-emb}
\end{figure}

The item kernel utilizes the dependency between the items in each time step. When learning the co-occurrence loss, the kernel hyperparameter and the item embedding captures the relational information of the co-occurrence. Figure \ref{item-emb}a illustrates the item embedding of movies. The item embedding with the same genres are distributed closely together. 

We generate the synthetic sequence to analyze the correlation from the trained kernel function. We use the counting and the item kernel combination without the user kernel because the sequence was synthetic. The synthetic sequence includes four different movie series and an animation movie. Figure \ref{item-emb}b shows that the movies belong to the same series have high correlations. 
On the contrary, the correlations between the animation genre and the other genres were low. 

\begin{figure}
	\includegraphics[width=\columnwidth]{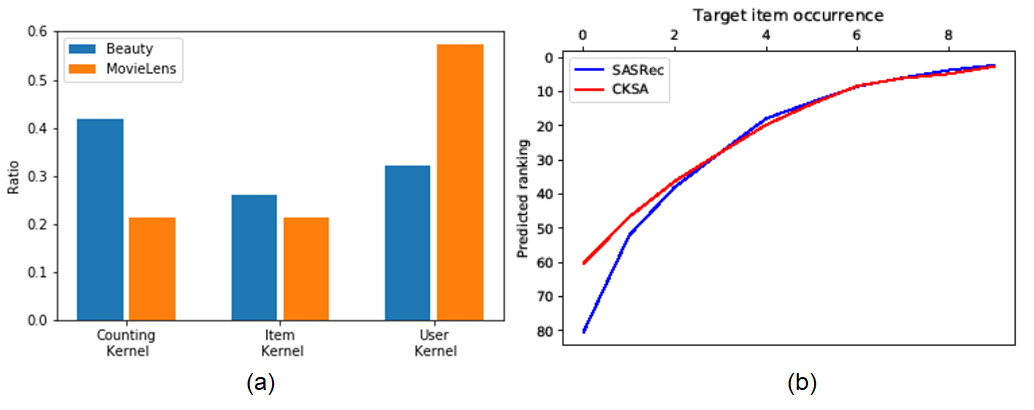}
	\caption{(a) The weights of the counting, the item, and the user kernels for the final kernel calculation (b) Average predicted ranking of the SASRec and RKSA by item occurrence in Beauty dataset. Value of the $x$-axis grows, it indicates the frequently occurred group. RKSA predict the higher ranking for infrequent items.}
	\label{infreq}
\end{figure}

Finally, we observed the weights of the counting, the item, and the user kernels, see Figure \ref{infreq}a, because the kernel weights also contribute to the construction of the correlation matrix. Since each dataset has different characteristics, a dataset emphasizes the counting, the item, and the user relations, differently. 
Interestingly, the counting kernel was not the most dominant kernel in MovieLens, but the user kernel was dominant. MovieLens is relatively dense dataset with respect to the number of average action per user, as shown in Table \ref{dataset}. Our proposed model, RKSA, adapts to the property of dataset well, and focus on the user kernel instead of other kernels on MovieLens dataset.

\subsubsection{Predicted ranking of infrequent items}
A sparse dataset, like Beauty, has many infrequent items, which are difficult to predict because of its information sparsity. To overcome this problem, RKSA utilizes the relational information of the whole dataset, instead of a single sequence in the prediction. Figure \ref{infreq}b shows that the target item is highly ranked by RKSA as the information sparsity worsens, compare to the predicted ranking of SASRec. 

\subsubsection{Attention Weight Case Study}
\begin{figure}
	\includegraphics[width=\columnwidth]{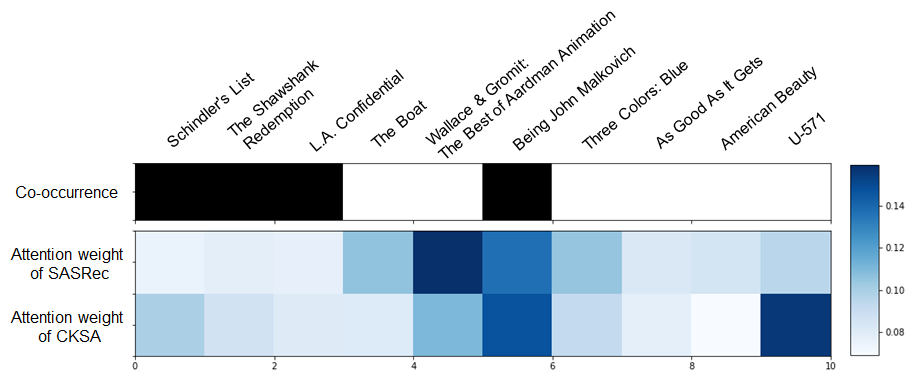}
	\caption{Attention heatmap for a user sequence of MovieLens. The first row indicates the co-occurrence, and the last item does not have co-occurrence information. If the co-occurrence between last item (query) and each item is bigger than average co-occurrence of sequence, we fill each timestep as black and the rest white. The second row is an attention weight in SASRec and the below is an attention weight in RKSA.}
	\label{fig:case}
\end{figure}

Figure \ref{fig:case} shows the attention weight of SASRec and RKSA with the co-occurrence information between the last item and each item of sequence. The sequence instance in Figure \ref{fig:case} has a high co-occurrence value at timestep 0, 1, 2, and 5; and Figure \ref{fig:case} confirms that RKSA places higher attention values than SASRec. In the opposite case, the attention weight of RKSA is lower than the attention weight of SASRec. 

\section{Conclusion}
We present relation-aware kernelized self-attention (RKSA) for a sequential recommendation task. RKSA introduces a new self-attention mechanism which is stochastic as well as kernelized by the relational information. While the past attention mechanisms are deterministic, we introduce a latent variable in the attention. Moreover, the latent variable utilizes the kernelized correlation matrix, so the kernel can be expanded to include relational information and modeling. From these innovations, we were able to see the best performance in all experimental settings. We expect that the further development on the stochastic attention of the Transformer will come in the near future.

\section{Acknowledgments}
This research was supported by Basic Science Research Program through the National Research Foundation of Korea(NRF) funded by the Ministry of Education (NRF-2018R1C1B6008652).

\bibliographystyle{aaai}
\bibliography{references}

\end{document}